# Secure Face Matching Using Fully Homomorphic Encryption

Vishnu Naresh Boddeti
Michigan State University
East Lansing, MI
vishnu@msu.edu

## Abstract

*Face recognition technology has demonstrated tremendous progress over the past few years, primarily due to advances in representation learning. As we witness the widespread adoption of these systems, it is imperative to consider the security of face representations. In this paper, we explore the practicality of using a fully homomorphic encryption based framework to secure a database of face templates. This framework is designed to preserve the privacy of users and prevent information leakage from the templates, while maintaining their utility through template matching directly in the encrypted domain. Additionally, we also explore a batching and dimensionality reduction scheme to trade-off face matching accuracy and computational complexity. Experiments on benchmark face datasets (LFW, IJB-A, IJB-B, CASIA) indicate that secure face matching can be practically feasible (16 KB template size and 0.01 sec per match pair for 512-dimensional features from SphereFace [23]) while exhibiting minimal loss in matching performance.*

## 1. Introduction

Face recognition is the science of determining the identity of people through their characteristic facial appearance. Being the primary modality for human identification, this topic has received tremendous attention in the pattern recognition and biometrics communities. There has been a steady improvement in face recognition accuracy over the past few decades, much of which can be attributed to deep learning over the last few years. Thanks to progress in learning feature representations, automated face recognition systems are now believed to surpass human performance in some scenarios [25]. As a result, face recognition technology is witnessing wide spread adoption in numerous applications, such as surveillance systems, social media and law enforcement. However, the increased deployment of such technology also makes it an attractive target of malicious attacks.

A typical face recognition system acquires a facial image, which undergoes some pre-processing, following which high-dimensional features are extracted. During enrollment, these features vectors are stored in a database along with their identity labels. This database is then used to verify a person's claimed identity (face verification) or determine a person's identity (face identification).

One of the simplest and foremost points of attack on a face recognition system is the database of face templates. Directly storing the facial feature vectors in the database can significantly compromise the privacy of the enrolled users and the security of the authentication system. For instance, it has been demonstrated recently that access to the representations can enable the reconstruction of a user's original facial image [26] by a malicious attacker. The reconstructed face images were shown to be of sufficient quality to be successfully matched by a state-of-the-art face recognition system. Similarly, soft facial attributes such as age, gender, ethnicity etc. can be predicted with high accuracy from facial features [24]. Therefore, it is imperative to devise techniques to prevent information leakage from face representations while preserving face matching performance, thereby preventing attacks of this nature. Addressing this problem is the central aim of this paper.

An attractive solution for secure face matching is the use of *cryptosystems* to protect the face template database and perform matching over encrypted data. These methods, however, have not found much traction in the biometrics and pattern recognition community. Generic encryption schemes are inherently incapable of supporting basic arithmetic operations, necessary for template matching, directly in the encrypted domain. On the other hand, *homomorphic cryptosystems* are a special class of cryptosystems that allow basic arithmetic operations over encrypted data [12]. These systems can be broadly categorized into three classes, (1) partially homomorphic encryption systems that support only a single arithmetic operation i.e., either addition [30] or multiplication [10] in the encrypted domain, (2) somewhat homomorphic encryption systems that support a limited number of both additions and multiplications, and (3) fully homomorphic encryption (FHE) systems that sup-port unlimited additions and multiplications directly in the encrypted domain. These systems offer a trade-off between generalization and computational complexity. Ergo, while the latter are more general they are significantly more computationally demanding than the former systems.

In this paper, we present a *fully homomorphic encryption* [13] based approach to (1) cryptographically secure the database of face templates and the feature vector of the presented face image, and (2) perform matching directly in the encrypted domain without the need for decrypting the templates. We leverage the observation that a typical face matching metric, either Euclidean distance or cosine similarity, can be decomposed into its constituent series of addition and multiplication operations. Both of these operations are supported by FHE schemes over encrypted data. This solution affords many benefits, including the desiderata of template protection schemes [27]:

**Diversity:** Using different encryption and decryption keys across different databases can protect against cross-database-matching attacks. **Revocability:** Compromised templates can be revoked and reissued simply by revoking and reissuing the encryption and decryption keys. **Security:** The security of the face template database is cryptographically guaranteed by the hardness of the ring learning with errors problem [6]. This problem forms the basis of current practical FHE schemes. Therefore, it is computationally prohibitive to obtain the original face template without the decryption key. **Performance:** Since homomorphic encryption admits addition and multiplication of integers directly in the encrypted domain, the proposed scheme can maintain the recognition performance, in terms of FAR and FRR, of the face recognition system. **Privacy:** The inability to access the original face template without the decryption keys helps preserve the privacy of the users, preventing estimation of either the original image [26] or any soft biometric traits [24].

The key technical barrier to realizing homomorphic encryption based face matching is the computational complexity of homomorphic encryption, especially homomorphic multiplication. This limitation renders it unsuitable for tasks that need to operate on high-dimensional vectors. For instance, we consider face representations that are hundreds to thousands of dimensions, thereby necessitating as many costly homomorphic operations. A straight-forward application of FHE for face matching needs 48.7 MB of memory for each 512-dimensional encrypted face template and 12.8 secs for matching a single pair of such templates [46].

The primary contributions of this paper are geared towards addressing this limitation: (1) Utilize a more efficient FHE scheme, namely Fan-Vercauteren [11], easing the computational burden to 16.5 MB of memory and 0.6 secs per pair of templates. These requirements could still be prohibitive for practical deployment. (2) Utilize a batching scheme that allows homomorphic multiplication of multiple values at the cost of a single homomorphic multiplication. This scheme reduces the computational requirements to, 16 KB memory per template and 0.01 secs to match a pair of templates. (2) Dimensionality reduction to further provide a trade-off between computational efficiency and matching performance.

## 2. Related Work

**Cryptographic Biometric Template Protection:** A number of approaches have been proposed to bring together biometric authentication and traditional cryptography. Soutar *et al.* [42, 43] proposed one of the earliest systems for biometric encryption of fingerprints. The idea of using visual cryptography [29] on raw biometric data was also explored in the context of fingerprint images [34], iris templates [36] and face images [37]. This method ensures biometric privacy while maintaining the ability to match biometric images. Fuzzy vault [19] based approaches have been proposed for biometric authentication in the context of fingerprint [47] and iris [22] patterns. These methods allow for matching inexact biometric templates in the encrypted domain. However, unlike the proposed FHE solution, they are not designed to maintain full utility (performance wise) of the encrypted biometrics.

**Pattern-Recognition Based Biometric Template Protection:** Instead of relying on the security afforded by cryptographic systems, a number of pattern recognition based approaches have been proposed for biometric template protection. Davida *et al.* [9] and Radha *et al.* [35] proposed non-invertible transformation functions to protect biometric signatures. Cancelable biometrics [32] are another approach for the same purpose while allowing for revocability and diversity of templates. Key-binding systems [28, 4] secure a biometric template by monolithically binding it with a secret key. The readers are encouraged to refer to the excellent article by Jain *et al.* [18] for a more comprehensive treatment of this topic. All the aforementioned approaches, however, trade-off security and matching performance i.e., they either suffer significant loss in matching accuracy or do not provide adequate template protection. In contrast, the FHE based scheme explored here does not suffer from this trade-off and can offer high levels of template protection without adversely sacrificing matching performance.

**Homomorphic Encryption for Biometrics:** The primary attraction of homomorphic encryption is the ability to perform basic arithmetic operations such as additions and multiplications in the encrypted domain. Initial homomorphic encryption [16] driven biometric authentication approaches [21, 3] were largely based on partial homomorphic encryption (PHE) schemes and in the context of matching binarized templates. Upmanyu *et al.* [48, 49] proposed a bio-metric matching system based on a PHE scheme. This scheme required repeated rounds of communication between the client and the database to compute the match score, suffering from large communication overhead and computational complexity. Barni *et al.* [1, 2] leverage a Paillier cryptosystem, a PHE system that supports addition over encrypted data. This system was used for matching fingerprints using fingercode templates where the probe is encrypted while the database remains unencrypted, providing no security to the fingerprint database. Penn *et al.* [33] also leverage a Paillier cryptosystem [30] for matching binary iris templates. Recently, Cheon *et al.* [7] proposed *Ghost-shell*, a secure biometric authentication scheme for matching 2048-bit binary iris templates in the encrypted domain. It relied on a somewhat homomorphic encryption scheme, requiring about 0.5s per match pair. However, these approaches are specific to matching binary bit strings and cannot maintain the same computational and algorithmic characteristics when extended to real valued face templates.

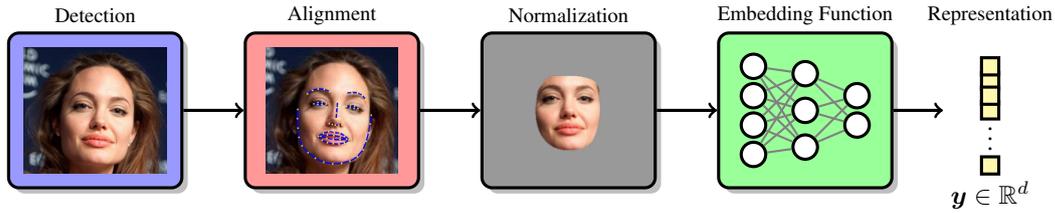

Figure 1: A typical face representation pipeline comprises of face detection, alignment, normalization and feature extraction. The embedding function maps a high-dimensional normalized face image to a $d$-dimensional vector representation.

The development of fully homomorphic encryption by Gentry *et al.* [15] has led to the promise of statistical analysis over encrypted data. In the context of biometric matching, Troncoso-Pastoriza *et al.* [46] proposed a FHE based face verification system using Gabor features. Their system was based on early an FHE scheme [14] that was computationally expensive (template size of 380 MB and 100 secs per match pair) and did not take full advantage of the specific nature of the face matching problem. In contrast, this paper proposes a secure face matching scheme that significantly improves the computational efficiency over encrypted templates (template size of 66 KB and 0.01 sec per match pair). We leverage the Fan-Vercauteren scheme [11], a more efficient FHE scheme, and a batching scheme that allows homomorphic multiplication of multiple numbers for the cost of a single homomorphic multiplication.

**Face Recognition:** There is a vast body of work devoted to studying the topic of face recognition. While an exhaustive survey of the literature is beyond the scope of this paper, we present a few approaches that form the basis of our face matching system. Current state-of-the-art face recognition approaches are based on deep neural network based representation learning techniques. These methods [44, 39, 31, 23] rely on end-to-end learning of feature extraction models using convolutional neural networks (CNNs). The goal of this paper is to ensure secure matching of face templates extracted from such representation models. We demonstrate the efficacy of our framework on two different face representation models, FaceNet [39] (128-dimensions) and SphereFace [23] (512-dimensions).

## 3. Approach

The goal of this paper is to explore the practicality of using fully homomorphic encryption for face matching over encrypted face templates. First we describe the face matching process and break it down to its constituent operations, series of multiplications and additions. Then we introduce fully homomorphic encryption and finally describe solutions to mitigate the computational burden of face matching over encrypted data.

### 3.1. Problem Setup

A typical face recognition system consists of a database of face representations $\boldsymbol{X} = \{\boldsymbol{x}_1, \ldots, \boldsymbol{x}_n\}$ extracted from the images of the enrolled subjects (see Fig. 1 for a pictorial illustration of the feature extraction process). During deployment, a user who wishes to get authenticated provides their facial image from which a representation $\boldsymbol{y}$ is extracted. This representation is then matched against the templates in the database. The result of the matching process is a score that determines the degree of dissimilarity between $\boldsymbol{y}$ and $\boldsymbol{X}$. The dissimilarity score between two representations $\boldsymbol{x} \in \mathbb{R}^d$ and $\boldsymbol{y} \in \mathbb{R}^d$ is typically defined as,

$$d(\boldsymbol{x}, \boldsymbol{y}) = 1 - \frac{\boldsymbol{x}^T \boldsymbol{y}}{\|\boldsymbol{x}\| \|\boldsymbol{y}\|} = 1 - \tilde{\boldsymbol{x}}^T \tilde{\boldsymbol{y}} = 1 - \sum_{i=1}^{d} \tilde{x}_i \tilde{y}_i \quad (1)$$

where $\tilde{\boldsymbol{x}} = \frac{\boldsymbol{x}}{\|\boldsymbol{x}\|}$ is the normalized representation. Hence, the face matching process is comprised of $d$ scalar multiplications and $d$ scalar additions.

In this work, we seek to devise a solution to preserve the privacy of the database $\boldsymbol{X}$ as well as the probe $\boldsymbol{y}$ and prevent leakage of any private information corresponding to a user's

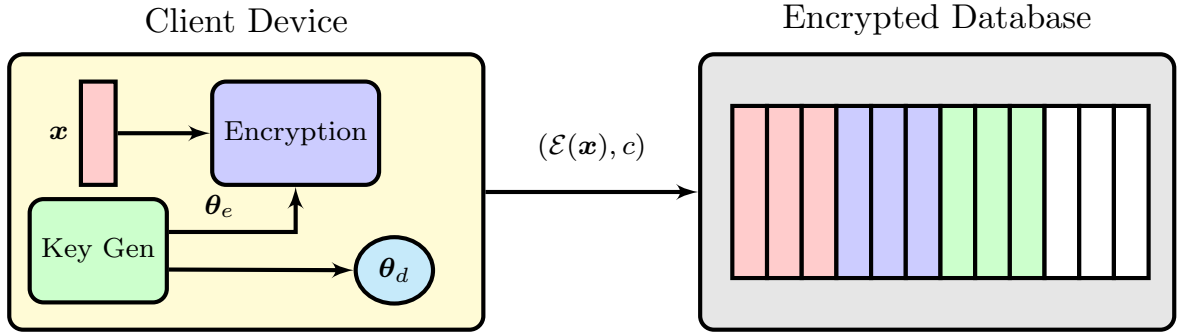

(a) **Enrollment:** During this phase the user generates the key pair $(\boldsymbol{\theta}_e, \boldsymbol{\theta}_d)$, encrypts feature vector $\boldsymbol{x}$ with the public key $\boldsymbol{\theta}_e$ and transmits the encrypted feature vector $\mathcal{E}(\boldsymbol{x})$ and the user identity $c$. The private key $\boldsymbol{\theta}_d$ however remains on the client device.

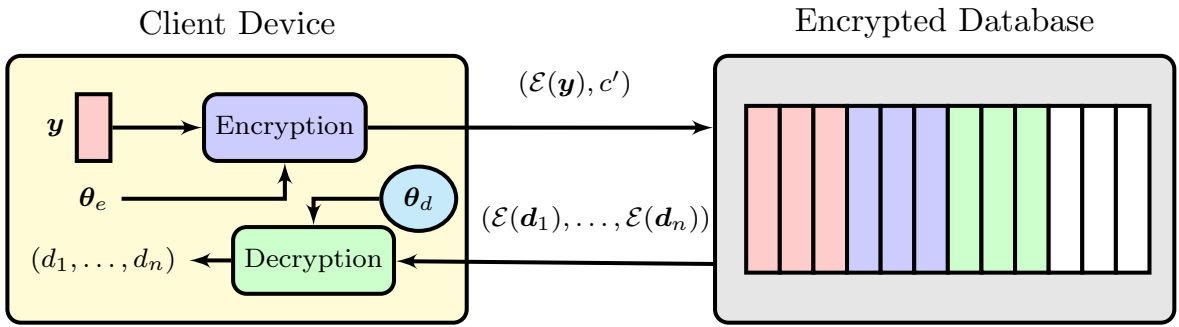

(b) **Authentication:** During this phase, a user seeks to get authenticated, encrypts the probe feature vector $\boldsymbol{y}$ with the public key $\boldsymbol{\theta}_e$ and transmits $\mathcal{E}(\boldsymbol{y})$ to the database along with the claimed identity $c'$. The database computes the dissimilarity score over encrypted features and transmits the matching score to the client, where it is decrypted with the private key $\boldsymbol{\theta}_d$.

Figure 2: Enrollment and authentication protocols for secure face matching using fully homomorphic encryption.

facial appearance. This can be achieved through a parameterized function that transforms a face template $\boldsymbol{z}$ from the original space into an alternate space, i.e., $\mathcal{E}(\boldsymbol{z}) = f(\boldsymbol{z}; \boldsymbol{\theta}_e)$, where $f(\cdot; \boldsymbol{\theta}_e)$ is the transformation function with parameters $\boldsymbol{\theta}_e$ and $\boldsymbol{z} = g(\mathcal{E}(\boldsymbol{z}); \boldsymbol{\theta}_d)$, where $g(\cdot; \boldsymbol{\theta}_d)$ is the inverse transformation function with parameters $\boldsymbol{\theta}_d$. While many such transformation functions exist, the key idea of our paper is to use transformation functions that can cryptographically guarantee the security of the original features $\boldsymbol{z}$ (without knowledge of $\boldsymbol{\theta}_d$) while allowing for template matching in the transformed space without loss of accuracy i.e.,

$$d(\mathcal{E}(\boldsymbol{x}), \mathcal{E}(\boldsymbol{y})) = d(f(\boldsymbol{x}, \boldsymbol{\theta}_e), f(\boldsymbol{y}, \boldsymbol{\theta}_e)) \approx d(\boldsymbol{x}, \boldsymbol{y}) \quad (2)$$

Transformation functions that satisfy this property enable us to preserve the privacy of the users. Even if a malicious attacker can gain access to the database of face templates, without access to the inverse transformation parameters $\boldsymbol{\theta}_d$ the attacker cannot estimate the user's facial image or extract soft biometric attributes inherent in the template.

### 3.2. Enrollment and Authentication Protocols

We now describe a protocol for user enrollment and authentication using the cryptographic transformation function described above. We consider a scenario with two entities, a *device* and a *database*.

**Enrollment:** During this phase the $c$-th user performs the following actions on the *device*; (1) generate pair of keys, a public encryption key $\boldsymbol{\theta}_e^c$ and a private decryption key $\boldsymbol{\theta}_d^c$ (each user can have their own pair of keys), (2) transmit the public key $\boldsymbol{\theta}_e^c$ to the *database*, (3) capture a series of images of the face, (4) extract the representation of the acquired face images $\boldsymbol{X}^c = \{\boldsymbol{x}_1^c, \ldots, \boldsymbol{x}_{n_c}^c\}$ where $n_c$ is the number of enrollment samples for the $c$-th user, (5) encrypt the face images using the public key $\boldsymbol{\theta}_e^c$, and (6) and finally transmit the encrypted templates to the database along with the identity label. On the *database* the encrypted templates, the public key $\boldsymbol{\theta}_e^c$ and identity label are added to the database. Figure 2a shows an illustration of this process.

**Authentication:** During this phase the *device* seeks to authenticate the user through the following process: (1) the user's facial image is acquired following which the representation $\boldsymbol{y}$ is extracted, (2) this representation is encrypted using the public key $\boldsymbol{\theta}_e$, which can possibly be different from $\boldsymbol{\theta}_e^c$, (3) transmit the encrypted representation, the claimed identity ($c'$) and the public encryption key $\boldsymbol{\theta}_e$ if it is different from $\boldsymbol{\theta}_e^c$. At the database if the encryption key of the probe is different from the encryption key of the user's template, the database executes key switching on the probe without having to decrypt it, (5) compute the dissimilarity score $d(\mathcal{E}(\boldsymbol{y}), \mathcal{E}(\boldsymbol{x})_k^{c'})$ for all the templates of the $c'$-th identity, (6) the encryption key of the scores are switched back to $\boldsymbol{\theta}_e$, and transmitted back to the device. Finally, at the device, the encrypted scores are decrypted using the secret decryption key $\boldsymbol{\theta}_d$ to verify the identity of the user. Figure 2b shows an illustration of this procedure.

### 3.3. Homomorphic Encryption: A Primer

We first introduce the main ideas behind the Fan-Vercauteren [11] FHE scheme that we use and describe a straight-forward but computationally expensive application of this scheme to the face matching problem.

**Mathematical Notation:** Modular arithmetic forms the mathematical basis of homomorphic encryption systems. For $t \in \mathbb{Z}$ a ring $R_t = \mathbb{Z}_t[x]/(x^n + 1)$ represents polynomials of degree less than $n$ with the coefficients modulo $t$. The operators $\lfloor \cdot \rfloor$, $\lceil \cdot \rceil$ and $\lfloor \cdot \rceil$ denote rounding down, up and to the nearest integer respectively. The operator $[\cdot]$ denotes the reduction of an integer by modulo $t$, where the reductions are performed on the symmetric interval $[-t/2, t/2)$. The operators when applied to a polynomial are assumed to act independently on the coefficients of the polynomial. $a \xleftarrow{\$} \mathcal{S}$ denotes that $a$ is sampled uniformly from the finite set $\mathcal{S}$. Similarly, $a \longleftarrow \chi$ denotes that $a$ is sampled from a discrete truncated Gaussian.

**Fan-Vercauteren Scheme [11]:** Any number, integer or rational, first needs to be encoded in a ring $R_t$ before it can be encrypted. Similarly, the encrypted numbers are encoded as polynomials in the ring $R_q$. Let $\lambda$ be the desired level of security, $w$ the base to represent numbers in, and $l = \lfloor \log_w q \rfloor$ the number of terms in the decomposition of $q$ into base $w$. The FV scheme utilizes three keys, (1) a private decryption key $\boldsymbol{\theta}_d$, (2) a public encryption key $\boldsymbol{\theta}_e$, and (3) evaluation keys $\boldsymbol{\theta}_{ev}$ which are necessary for multiplication over encrypted data. Below are the details of key generation, addition and multiplication over encrypted integers.

- Gen Secret Key ($\lambda$): Sample $\boldsymbol{\theta}_d \xleftarrow{\$} R_2$

- Gen Public Key ($\boldsymbol{\theta}_d$): Sample $\boldsymbol{a} \xleftarrow{\$} R_q$ and $\boldsymbol{e} \xleftarrow{\$} \chi$. Output $\boldsymbol{\theta}_e = ([-(\boldsymbol{a}\boldsymbol{\theta}_d + \boldsymbol{e})]_q, \boldsymbol{a})$

- Gen Evaluation Key ($\boldsymbol{\theta}_d, w$): for $i \in \{0, \ldots, l\}$, sample $\boldsymbol{a}_i \xleftarrow{\$} R_q, \boldsymbol{e}_i \longleftarrow \chi$.
  Output $\boldsymbol{\theta}_{ev} = ([-(\boldsymbol{a}_i \boldsymbol{\theta}_d + \boldsymbol{e}_i) + w^i \boldsymbol{\theta}_d^2]_q, \boldsymbol{a}_i)$

- Encrypt($\boldsymbol{m}, \boldsymbol{\theta}_e$): For $\boldsymbol{m} \in R_t$, sample $\boldsymbol{u} \xleftarrow{\$} R_2$ and $\boldsymbol{e}_1, \boldsymbol{e}_2 \longleftarrow \chi$. Output $\boldsymbol{ct} = ([\boldsymbol{m} + \boldsymbol{\theta}_e[0]\boldsymbol{u} + \boldsymbol{e}]_q, [\boldsymbol{\theta}_e[1]\boldsymbol{u} + \boldsymbol{e}]_q)$

- Decrypt($\boldsymbol{ct}, \boldsymbol{\theta}_d$): Output $\left[ \left\lfloor \frac{t}{q} \left[ \boldsymbol{ct}[0] + \boldsymbol{ct}[1]\boldsymbol{\theta}_d \right] \right\rceil \right]$

- Add($\boldsymbol{ct}_0, \boldsymbol{ct}_1$): Output $(\boldsymbol{ct}_0[0] + \boldsymbol{ct}_1[0], \boldsymbol{ct}_0[1] + \boldsymbol{ct}_1[1])$

- Multiply($\boldsymbol{ct}_0, \boldsymbol{ct}_1$): Compute:
  $\boldsymbol{c}_0 = \left[ \left\lfloor \frac{t}{q} (\boldsymbol{ct}_0[0]\boldsymbol{ct}_1[0]) \right\rceil \right]_q$
  $\boldsymbol{c}_1 = \left[ \left\lfloor \frac{t}{q} (\boldsymbol{ct}_0[0]\boldsymbol{ct}_1[1] + \boldsymbol{ct}_0[1]\boldsymbol{ct}_1[0]) \right\rceil \right]_q$
  $\boldsymbol{c}_2 = \left[ \left\lfloor \frac{t}{q} (\boldsymbol{ct}_0[1]\boldsymbol{ct}_1[1]) \right\rceil \right]_q$
  $\boldsymbol{c}_2 = \sum_{i=0}^{l} c_2^{(i)} w^i$ and compute
  $\boldsymbol{c}'_0 = \boldsymbol{c}_0 + \sum_{i=1}^{l} \boldsymbol{\theta}_{ev}[i][0]c_2^i$
  $\boldsymbol{c}'_1 = \boldsymbol{c}_1 + \sum_{i=1}^{l} \boldsymbol{\theta}_{ev}[i][1]c_2^i$
  Output $(\boldsymbol{c}'_0, \boldsymbol{c}'_1)$

The algorithm described above enables us to perform face matching in the encrypted domain. The face dissimilarity score described in Eq. 1 can be computed in the encrypted domain as:

$$d(\boldsymbol{x}, \boldsymbol{y}) = 1 - g\left( \sum_{i=1}^{d} \mathcal{E}(\boldsymbol{x}_i)\mathcal{E}(\boldsymbol{y}_i), \boldsymbol{\theta}_d \right) \quad (3)$$

where computing the inner product over encrypted data would necessitate $d$ encrypted multiplications and $d - 1$ encrypted additions. Since each encrypted multiplication is computationally expensive, evaluating the face dissimilarity score can be very slow [46]. This problem is further exacerbated by the high dimensionality of the face templates, typically running into hundreds to thousands of dimensions.

Addition and multiplication of two encrypted integers translates to polynomial addition and multiplication in $R_t$. As long as the coefficients of the resulting polynomials do not exceed $t$ and their degrees do not exceed $(x^n + 1)$, correctness is ensured and we can recover the score by evaluating the polynomial at the integer base $w$ i.e., at $x = w$.

### 3.4. Efficient Homomorphic Face Matching

While the algorithm described above can be directly utilized for face matching, it suffers from high computational requirements. Supporting computations involving many high-precision real-values from the facial features requires prohibitively large values of $n$ and $t$. For a 512-dimensional face template and desired security level of 128 bits, using this algorithm requires 16.5 MB of storage space for each encrypted template and 0.6 secs for matching a pair of such templates. We present a few different solutions to address these limitations. First, we describe a quantization scheme to encode facial features in $R_t$. Then, we describe a batching scheme for encoding a vector of real-values into a single polynomial using the Chinese Remainder Theorem. This encoding allows us to amortize multiple homomorphic multiplications within a single homomorphic multiplication, thereby providing significant computational efficiencies. Lastly, we leverage classical dimensionality reduction algorithms to further ease the computational burden of homomorphic face matching.

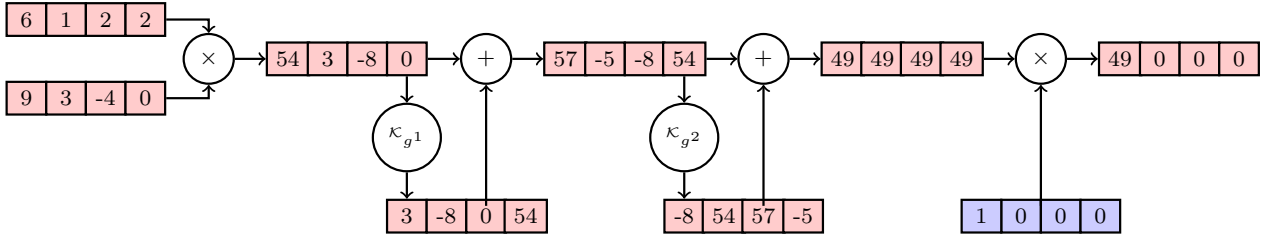

Figure 3: Homomorphic computation of inner product between vectors using batching, where multiple elements are encrypted into a single encrypted block. The batching process does not allow access to each element of the encrypted block. Therefore, after the Hadamard product, the sum of the elements within the encrypted block can be computed through repeated cyclic rotations and additions.

**Data Representation:** The efficiency of homomorphic additions and multiplications is critically dependent on the encoding scheme chosen to represent the real-valued elements of the face template within a polynomial ring $R_t$. The space of ring $R_t$ is significantly different from the space of real values $\mathbb{R}$ or the space of integers $\mathbb{Z}$. Hence, to maintain the integrity of the homomorphic computations, the chosen encoding scheme needs to ensure that the range of values after the desired homomorphic operations remain within the ring. Pursuantly, we encode the real-valued face features into integers with a precision of two decimal digits and then represent these integers in base $w$.

$$Encoding(a) = sign(a)(a_{n-1}x^{n-1}+\cdots+a_1x+a_0) \quad (4)$$

Accordingly, this encoding scheme can represent any integer in $-(w^n - 1) \leq a \leq (w^n - 1)$.

**Batching:** The main computational bottleneck of face matching over encrypted data is the number of homomorphic multiplications necessary for computing the dissimilarity score. Brakerski *et al.* [5] and Smart *et al.* [41] proposed techniques for homomorphic encryption and decryption over an array of numbers instead of a single number at a time. These techniques leverage the Chinese Remainder Theorem (CRT) and encode multiple numbers within the same polynomial. For instance, given the primitive $2n$-th root of unity modulo $t$, we have: $x^n + 1 = (x - \xi)(x - \xi^3)\ldots(x - \xi^{2n-1}) \mod t$

The CRT factorizes the ring $R_t$ when $t$ is chosen to be a product of many small prime factors i.e., $t = \prod_{i=1}^{k} t_i$, where $p_1,\ldots,p_k$ are distinct prime numbers. Given these primes, we can perform computations modulo each prime and recover the computation with respect to the product of the primes by using the CRT. The computations with respect to the primes are independent of each other and hence can be implemented in parallel.

Given two vectors $\boldsymbol{x}$ and $\boldsymbol{y}$, the sum and Hadamard product of the two vectors can simply be computed as element-wise addition and multiplication, respectively, of the encrypted vectors. By batching $k$ numbers into a single encrypted block, we can perform $k$ homomorphic additions or multiplications at the block level for the computational cost of a single operation. Therefore, as the batch size gets larger, we can realize significant computational efficiency. The main drawback of this scheme, however, is the inability to access the individual elements of the packed encrypted vector. This prevents us from computing the sum of the elements, an operation that is necessary for computing the dissimilarity score (Eq. 3). This limitation can be addressed by leveraging the observation made my Gentry *et al.* [15], namely, it is possible to cyclically rotate the encrypted vectors without the need for decryption. Therefore, the sum of the encrypted values can be computed by cyclically rotating and adding the encrypted vectors $l = \log_w^q$ times. Figure 3 and Algorithm 1 respectively provide a pictorial illustration and algorithmic description of this process.

---

**Algorithm 1** Homomorphic Inner Product

1: **procedure** ENCRYPTEDINNERPRODUCT($\boldsymbol{m}, \boldsymbol{m'}$)
2:     $\mathcal{E}(\boldsymbol{x}) \leftarrow f(\boldsymbol{x}, \boldsymbol{\theta}_e)$     ▷ CRT Encoding
3:     $\mathcal{E}(\boldsymbol{y}) \leftarrow f(\boldsymbol{y}, \boldsymbol{\theta}_e)$     ▷ CRT Encoding
4:     $\boldsymbol{ct} \leftarrow Multiply(\mathcal{E}(\boldsymbol{x}), \mathcal{E}(\boldsymbol{y}))$   ▷ Batch Multiply
5:     **for** $i = \{0,\ldots,l\}$ **do**
6:         $\boldsymbol{ct} \leftarrow \boldsymbol{ct} + \mathcal{K}_{g^i}(\boldsymbol{ct})$   ▷ Cyclic Rotation
7:     **end for**
8:     **return** $\boldsymbol{ct}$   ▷ The element in the first slot is the desired inner product.
9: **end procedure**

---

**Dimensionality Reduction:** The high computational cost of face matching over encrypted data depends on two factors, (1) the high computational cost of one homomorphic multiplication, and (2) the high dimensionality of the face templates. While the former is addressed to an extent by the batching scheme, there are further avenues for mitigating the computational burden by reducing the dimensionality of the face template. While a variety of dimensionality reduction algorithms exist (ISOMAP [45], LLE [38], random projections [8] etc.), we consider classical Principal Component Analysis based dimensionality reduction as an illustrative example.

Table 1: Security Parameters, Timing and Memory

| Security in bits ($\lambda$) | Dim ($d$) | No FHE | | Parameters | | | Batching | | | | | No Batching | | | | |
|---|---|---|---|---|---|---|---|---|---|---|---|---|---|---|---|---|
| | | Time ($\mu s$) | Mem (KB) | $n$ | $t$ (bits) | $q$ | Time (ms) | | | | Mem (KB) | Time (ms) | | | | Mem (MB) |
| | | | | | | | Enc | Score | Dec | Total | | Enc | Score | Dec | Total | |
| 128 | 64 | 0.44 | 2.0 | 128 | 110 | 40961 | 0.07 | 0.17 | 0.01 | 0.25 | 2.0 | 4.40 | 5.25 | 0.01 | 9.66 | 0.25 |
| | 128 | 0.89 | 4.0 | 256 | 110 | 40961 | 0.14 | 0.38 | 0.02 | 0.59 | 4.0 | 17.57 | 21.05 | 0.02 | 38.64 | 1.0 |
| | 512 | 3.48 | 16.0 | 1024 | 110 | 40961 | 0.58 | 1.80 | 0.07 | 2.45 | 16.0 | 280.19 | 343.81 | 0.08 | 624.07 | 16.5 |
| | 1024 | 7.49 | 32.0 | 2048 | 110 | 40961 | 1.14 | 4.02 | 0.15 | 5.80 | 33.0 | 1135.44 | 1411.82 | 0.16 | 2547.42 | 66.0 |
| | 1024 | 7.49 | 32.0 | 4096 | 110 | 40961 | 2.27 | 8.36 | 0.30 | 11.42 | 66.0 | 2214.88 | 2924.75 | 0.33 | 5139.97 | 131.0 |
| 192 | 64 | 0.44 | 2.0 | 128 | 77 | 40961 | 0.07 | 0.17 | 0.01 | 0.25 | 2.0 | 4.38 | 5.25 | 0.01 | 9.64 | 0.25 |
| | 128 | 0.89 | 4.0 | 256 | 77 | 40961 | 0.15 | 0.39 | 0.02 | 0.55 | 4.0 | 17.24 | 21.07 | 0.02 | 38.33 | 1.0 |
| | 512 | 3.48 | 16.0 | 1024 | 77 | 40961 | 0.58 | 1.85 | 0.07 | 2.50 | 16.0 | 274.11 | 343.60 | 0.08 | 617.79 | 16.5 |
| | 1024 | 7.49 | 32.0 | 2048 | 77 | 40961 | 1.17 | 4.03 | 0.15 | 5.35 | 33.0 | 1134.22 | 1410.99 | 0.16 | 2545.38 | 66.0 |
| | 1024 | 7.49 | 32.0 | 4096 | 77 | 40961 | 2.33 | 8.91 | 0.30 | 11.54 | 66.0 | 2214.96 | 2889.39 | 0.33 | 5104.68 | 131.0 |

**Key Switching:** In a practical deployment of face recognition systems, it is natural to consider the scenario where each user utilizes their own set of public encryption ($\theta_e^c$) and private decryption ($\theta_d^c$) keys respectively. This results in a scenario where face templates encrypted through one set of encryption keys have to be re-encrypted through another set of encryption keys without decrypting the template. This key-switching operation is supported by the FV scheme and is similar to the multiplication operation in Section 3.3.

### 3.5. Security Analysis:

The proposed system has three sources of attack, (1) the client device, (2) the communication channel between the client and the database, and (3) the database itself. The security of the client device is of paramount importance, since the device stores the private decryption key and is also the source of the raw facial images and the extracted facial features. We assume the "honest-but-curious" security model where all the parties honestly follow the protocol and learn nothing beyond their own outputs. Under this security model, the communication channel is secure from a malicious attacker, since the encrypted facial template cannot be decrypted without access to the secret key. Similarly the facial database itself, which consists of the user's identity label and encrypted facial templates, is secure without the access to the secret decryption keys.

## 4. Experiments

The goal of this paper is to explore the practicality and efficacy of using fully homomorphic encryption for the purpose of face matching over encrypted data. For our experimentation, we consider two different state-of-the-art deep neural network based face representations, namely, FaceNet [39] and SphereFace [23] with 128 and 512 (or 1024) dimensional features, respectively.

We evaluate these representations over multiple datasets of varying complexity. Specifically, we consider the following datasets (1) **LFW [17]** consisting of 13,233 face images of 5,749 subjects, downloaded from the web. These images exhibit limited variations in pose, illumination, and expression, (2) **IJB-A [20]:** The IARPA Janus Benchmark-A (IJB-A) dataset consisting of 500 subjects with a total of 25,813 images (5,399 still images and 20,414 video frames). This is a challenging dataset exhibiting i) full pose variation, ii) a mix of images and videos, and iii) wider geographical variation of subjects, (3) **IJB-B [50]:** The IARPA Janus Benchmark-B (IJB-B) dataset consists of 1,845 subjects with a total of 76,824 images (21,798 still images and 55,026 video frames from 7,011 videos). This dataset is a larger superset of the IJB-A dataset, with more subjects and greater facial appearance variations, and (4) **CASIA [51]:** A large collection of labeled images downloaded from the web (based on names of famous personalities) typically used for training deep neural networks. It consists of 494,414 images across 10,575 subjects, with an average of about 500 face images per subject. This dataset was used to train the FaceNet and SphereFace representations.

### 4.1. Implementation Details

Our implementation is based on the NTL C++ Library [40]. For all experiments, we used a machine with a 4-core Intel i5-6400 processor running at 2.7 GHz. All of our experiments were run in a single-threaded environment. Before we perform face matching experiments it is crucial to determine the parameters necessary for security and correctness of the face matching process. For a given desired level of security, Table 1 shows the effect of the different parameter settings ($n$, $q$ and $t$) and the dimensionality of the facial features on the computational complexity of face matching over encrypted features. We report a breakdown of the time taken for each of the steps in the matching process, the total matching time and the size of the templates. We observe that the fastest FHE based face matching is admittedly over 1000x slower than matching in the real-domain. Nevertheless, batching can offer significant a speed-up ($T$=0.59ms vs $T$=38.64ms for n=256, d=128 and $T$=11.42ms vs $T$=624ms for n=1024, d=512) and memory savings (4 KB vs 1 MB for n=256, d=128, 16 KB vs 16.5 MB for n=1024, d=512) over element-wise homomorphic face matching. In comparison, the FHE based face matching scheme by Troncoso-Pastoriza *et al.* [46] requires ∼12.8 secs per match pair and 48.7 MB for each template for 512-dimensional features. In contrast, our results, suggest that face matching over encrypted templates using the proposed FHE framework can provide high levels of security, 128 to 192 bits, and real-time matching over a small database of face templates.

Table 2: Face Recognition Accuracy (TAR @ FAR in %)

| Dataset | Method | 128-D FaceNet | | | 512-D SphereFace | | | 64-D PCA FaceNet | | | 64-D PCA SphereFace | | |
|---|---|---|---|---|---|---|---|---|---|---|---|---|---|
| | | 0.01% | 0.1% | 1% | 0.01% | 0.1% | 1% | 0.01% | 0.1% | 1% | 0.01% | 0.1% | 1% |
| LFW | No FHE | 84.06 | 94.56 | 98.65 | 90.49 | 96.74 | 99.11 | 83.99 | 94.64 | 98.73 | 88.41 | 95.80 | 98.87 |
| | FHE ($2.5\times10^{-3}$) | 84.05 | 94.56 | 98.65 | 90.49 | 96.74 | 99.11 | 84.00 | 94.64 | 98.72 | 88.38 | 95.80 | 98.87 |
| | FHE ($1.0\times10^{-2}$) | 83.89 | 94.53 | 98.66 | 90.46 | 96.73 | 99.11 | 83.94 | 94.62 | 98.72 | 88.38 | 95.80 | 98.87 |
| | FHE ($1.0\times10^{-1}$) | 78.82 | 92.58 | 98.35 | 84.78 | 94.72 | 98.64 | 78.18 | 93.06 | 98.41 | 84.32 | 94.38 | 98.69 |
| IJB-A | No FHE | 23.13 | 45.92 | 70.26 | 7.27 | 30.40 | 67.89 | 20.13 | 46.07 | 73.71 | 6.81 | 28.92 | 67.59 |
| | FHE ($2.5\times10^{-3}$) | 23.12 | 45.94 | 70.27 | 7.27 | 30.40 | 67.89 | 20.08 | 46.14 | 73.66 | 6.81 | 28.92 | 67.59 |
| | FHE ($1.0\times10^{-2}$) | 22.83 | 45.78 | 70.17 | 7.25 | 30.39 | 67.66 | 19.92 | 46.05 | 73.81 | 6.80 | 28.92 | 67.45 |
| | FHE ($1.0\times10^{-1}$) | 19.82 | 42.74 | 68.37 | 7.20 | 30.00 | 65.23 | 17.54 | 44.02 | 72.11 | 6.68 | 27.43 | 66.76 |
| IJB-B | No FHE | 25.77 | 48.31 | 74.47 | 7.86 | 31.27 | 69.83 | 24.95 | 47.80 | 74.58 | 7.13 | 29.72 | 69.79 |
| | FHE ($2.5\times10^{-3}$) | 25.78 | 48.28 | 74.46 | 7.86 | 31.27 | 69.82 | 25.06 | 47.78 | 74.66 | 6.85 | 29.70 | 69.69 |
| | FHE ($1.0\times10^{-2}$) | 25.71 | 48.31 | 74.44 | 7.80 | 31.29 | 69.75 | 24.90 | 47.97 | 74.50 | 6.86 | 29.85 | 69.52 |
| | FHE ($1.0\times10^{-1}$) | 23.75 | 46.08 | 72.87 | 7.49 | 30.92 | 67.45 | 22.79 | 45.80 | 73.12 | 6.90 | 29.82 | 69.55 |
| CASIA | No FHE | 70.98 | 84.70 | 93.29 | 86.48 | 90.81 | 93.83 | 70.85 | 84.76 | 93.35 | 82.87 | 89.07 | 92.80 |
| | FHE ($2.5\times10^{-3}$) | 70.98 | 84.70 | 93.29 | 86.47 | 90.81 | 93.83 | 70.94 | 84.82 | 93.33 | 82.78 | 89.04 | 92.78 |
| | FHE ($1.0\times10^{-2}$) | 70.96 | 84.68 | 93.28 | 86.47 | 90.83 | 93.81 | 70.92 | 84.82 | 93.32 | 82.74 | 89.00 | 92.79 |
| | FHE ($1.0\times10^{-1}$) | 70.82 | 84.66 | 93.25 | 83.21 | 89.49 | 93.05 | 70.88 | 84.79 | 93.32 | 78.82 | 87.29 | 92.16 |

## 4.2. Homomorphic Face Matching

We report the results of face matching experiments on benchmark datasets for two different, state-of-the-art face representations (FaceNet [39] and SphereFace [23]) in Table 2. For each of these representations, aligned and normalized face images are propagated through the corresponding neural networks to extract[1] facial templates of 128 and 512 dimensions, respectively. We report true acceptance rate (TAR) at three different operating points of 0.01%, 0.1% and 1.0% false accept rates (FARs). We first evaluate face matching performance over unencrypted face templates as a baseline to compare against. Since FHE operates on integer representations of the face templates, we consider three different quantization schemes for each element of the face features, one with a precision of 0.1, 0.01 and another with a precision of 0.0025. We observe that the performance of the latter scheme, by way of being a more precise representation of the raw features, is competitive with the performance of the unencrypted features. As a result, the homomorphic face matching over encrypted features can perform as well as matching raw features while providing template protection, preventing information leakage and preserving the privacy of the users. Finally, our experimental results suggest that even simple PCA based dimensionality reduction can perform comparably to the original high-dimensional features, while affording significant homomorphic face matching efficiency.

## 5. Conclusion

This paper explored the practicality of securing and matching a database of face templates using fully homomorphic encryption. We utilize a batching scheme, that performs multiple homomorphic multiplications in a single operation, and a dimensionality reduction scheme. Our framework affords significant computational efficiencies, requiring 16 KB of memory space per encrypted template and 0.01 secs for matching a pair of such templates. Experimental results over multiple benchmark datasets suggest that, fully homomorphic encryption could be a viable practical solution for accurate face matching in the encrypted domain, and can help prevent information leakage and preserve the privacy of users.

---

[1]We use publicly available implementations, https://github.com/davidsandberg/facenet and https://github.com/wy1iu/sphereface.